\documentclass[letterpaper, 10 pt, conference]{ieeeconf}  

\IEEEoverridecommandlockouts                              

\usepackage{graphicx} 
\usepackage{amsmath} 
\usepackage{amssymb}  
\usepackage{gensymb}
\usepackage{algorithm}
\usepackage[noend]{algpseudocode}
\usepackage{ctable}
\usepackage{subcaption}
\usepackage{hyperref}

\title{\LARGE \bf
Resilient Sensor Architecture Design and Tradespace Analysis for Autonomous Vehicle Localization and Mapping
}

\author{Anne Collin, Antonio Ter\'an Espinoza
\thanks{This work was partially supported by Hitachi Autonomotive Systems America, Inc., and CONACyT, M\'exico.}
\thanks{All authors are with the Department of Aeronautics and Astronautics, Massachusetts Institute of Technology, Cambridge, MA 02139 USA
        {\tt\small \{acollin,teran\}@mit.edu}}%
 }

\begin{document}

\maketitle
\thispagestyle{empty}
\pagestyle{empty}

\begin{abstract}

  As autonomous cars are rolled out into new environments, their ability to  solve the simultaneous localization and mapping (SLAM) problem becomes  critical. In order to tackle this problem, autonomous vehicles rely on sensor suites that provide them with information about their operating environment. When large scale production is taken into consideration, a trade-off between an acceptable sensor suite cost and its resulting performance characteristics arises. Furthermore, guaranteeing the system's performance requires a resilient sensor network design. This work seeks to address such trade-offs by introducing a method that takes into account the performance, cost, and resiliency of distinct sensor selections. As a result, this method is able to offer sensor combination recommendations based on the vehicle's operating environment. It is found that the structure of the environment influences sensor placement, and that the design of a resilient sensor network involves careful consideration of both environmental attributes such as landmark density and location, as well as the available types of complimentary sensors. Demonstration of the proposed approach is shown by evaluating it using sequences from the KITTI Benchmark Suite.

\end{abstract}

\section{Introduction and Related Work}

In recent years, research and development efforts around autonomous cars have shifted towards producing certifiable systems capable of being deployed into multiple environments~\cite{Junietz2018}. Naturally, sensors play a primordial role within this system by enabling the vehicle to interact with and to account for the ever-changing operational scenarios. Unfortunately, the plethora of available sensor types~\cite{Chong2015}, in addition to their possible multiplicity and physical location within the vehicle's system, lead to a combinatorial number of possible sensor suite designs from which to choose. This poses both a design and a testing problem, since it becomes monetarily infeasible to iterate through all possible physical prototypes and computationally intractable to account for every combination in driving simulations~\cite{Broggi2013}.

Amongst the most important metrics to consider when choosing an appropriate set of sensors with which to outfit an autonomous vehicle are performance~\cite{Ulbrich2017}, reliability~\cite{ISO26262}, and cost~\cite{Bosch2018}. The quality of the SLAM solution has a direct impact on the behavior of the car, since its output is utilized by the planning and control module for essential guidance tasks~\cite{Ulbrich2017}. In addition to preserving the warranted performance requirements, the sensor suite of choice must be resilient; it needs to guarantee safe operations in new and unknown environments~\cite{Ort2018}, as well as robustness against hardware failures and loss of measurements (e.g., unavailable GPS information in certain urban settings)~\cite{Jo2015}. Moreover, for autonomous cars to become a viable solution, they need to satisfactorily reach the aforementioned system-level performance while maintaining competitive costs in order to successfully rival traditional alternative transportation systems (e.g., private autos, public transportation)~\cite{Wadud2017, Bosch2018}.

Recent advances in smoothing techniques and optimization of submodular functions have encouraged the development of design methods to improve SLAM solutions, resulting in task-specific approaches geared towards resource-constrained systems. It has been shown that for a given sensor budget, optimization over an LQG-control performance metric yields a method to consciously choose which sensors to toggle during real-time operations to reduce uncertainty~\cite{Tzoumas2017}. On the other hand, alternative routes have opted to work directly with sensor information; by optimizing for accuracy when selecting features for visual odometry~\cite{Zhao2018} or by exploiting notions such as anticipation to prune irrelevant visual cues in visual-inertial navigation~\cite{carlone2018attention}, performance can be targeted.

However, to the knowledge of the authors, there exists a gap within the application space of such methods to help guide and inform the sensor choice and placement on autonomous vehicles, and to aid in the quantification of the tradespace between performance and non-uniform sensor costs. Initial efforts that make use of similar techniques have been presented~\cite{Collin2018}, but they still lack a systematic strategy to tackle the sensor combination enumeration. Furthermore, none of these advances consider the resiliency of sensor networks.

This work presents a method to concurrently address the multiple trade-offs posed by autonomous vehicle's system requirements through the use of greedy enumeration strategies paired with submodularity concepts. The {\bf main contributions} of the proposed sensor selection strategy are answering the following: {\bf i)} how to best address the sensor selection problem while satisfying the system's constraints; {\bf ii)} how to develop a quantitative mapping of the relevant tradespace; and {\bf iii)} how to introduce the concept of resiliency into the selection process.

The paper comprises the following sections: Section~\ref{sec:background} provides the insight and motivation behind the chosen inference methods and algorithms; the problem statement and formulation is presented in Section~\ref{sec:problem}; Section~\ref{sec:methodology} describes the proposed approach; experimental results are shown in Sections~\ref{sec:tradespace} and~\ref{sec:resilience}; concluding remarks are presented in Section~\ref{sec:conclusion}.

\section{Background Information and Preliminaries}\label{sec:background}

SLAM systems are typically broken down into a back-end component in charge of inference, and a front-end component that handles the perception and data association aspect of the problem~\cite{Dellaert2012}. We therefore do not consider the increased difficulty in data association with certain sensor types.
This work focuses on the inference aspect of SLAM, and how a solution is affected by the information that is chosen. Factor graphs are chosen as the framework with which to carry out such inferences.

Factor graphs are undirected graphical models that represent joint probability distributions in such a way that conditional independencies can be leveraged to exploit the inherent sparsity of a problem~\cite{Dellaert2012}. 


It is well known how the nonlinear least squares formulation that emerges from the factor graph modeling technique can be transfomed, through a series of linearizations and whitening processes, into the standard least squares problem~\cite{kaess2008isam}
\begin{equation}\label{eq:stdLS}
  \boldsymbol{\theta}^* = \arg\min_{\boldsymbol{\theta}}|| \mathbf{A}\boldsymbol{\theta} - \mathbf{b} ||_2^2,
\end{equation}
\noindent where $\boldsymbol{\theta} \in \mathbb{R}^{n}$ is the vector containing all pose and landmark variables of interest from $\mathbf{X}$, the set of all vehicles states, and $\mathbf{L}$, the set of all landmarks, $\mathbf{A} \in \mathbb{R}^{m\times n}$ is the sparse measurement Jacobian with $m$ measurement rows, and $\mathbf{b} \in \mathbb{R}^m$ the right-hand side vector. Furthermore, a QR factorization of the matrix $\mathbf{A} = \mathbf{Q}\; [ \mathbf{R} \;\; \mathbf{0} ]^\top$ can be performed in order to extract the system's square root information matrix $\sqrt{\boldsymbol{\Lambda}} = \mathbf{R} \in \mathbb{R}^{n\times n}$, which allows for the efficient computation of the SLAM solution through simple back-substitution as $\mathbf{R} \boldsymbol{\theta}^* = \mathbf{d}$, with $[\mathbf{d}\;\;\mathbf{e}]^\top = \mathbf{Q}^\top \mathbf{b}$~\cite{Kaess2009}.

Factor graph models constitute a resource for sensor combination performance evaluation. As far as sensor enumeration is concerned, choosing a set of sensors that maximizes a sensing objective, such as the log-determinant of the information matrix, under cardinality constraints, is usually NP-hard \cite{Krause2014}.
Solutions include solving a convex relaxation of the problem \cite{joshi2009sensor}, or leveraging the diminishing returns property, called submodularity, of the objective function to design a greedy algorithm, guiding the search through the design space \cite{Krause2008}.

Several performance criteria can be chosen to evaluate the quality of the SLAM estimate. The log determinant of the covariance matrix, which represents the volume of the uncertainty ellipse around the estimate \cite{joshi2009sensor}, is selected for this work. It has been proven to be a valid evaluation metric for SLAM \cite{Carrillo2012}. Additionally, with the covariance matrix $\boldsymbol{\Sigma} = \boldsymbol{\Lambda}^{-1}$, and since the R matrix is upper triangular, we have 
\begin{align}
\log \det(\mathbf{\Lambda}) =  & \log \det(\mathbf{A^{T}A}) \\
= & \log \det(\mathbf{R^{T}R}) \\
= & 2 \sum_i \log(R_{ii})
\end{align}
Thus, this evaluation does not impose a computationally prohibitive overhead to the SLAM problem resolution.




\section{Problem Formulation and Statement}\label{sec:problem}

The available pool of sensors with which to equip an autonomous vehicle can be expressed as a set $\mathcal{S}$. The subset $\mathbf{s} \subseteq \mathcal{S}$ is denoted as the vehicle's current sensor suite, which gathers a collection of measurements $\mathbf{Z}$ as it interacts with its environment. From Section~\ref{sec:background}, it is clear how the measurement Jacobian matrix $\mathbf{A}$ is a direct function of the set of acquired measurements $\mathbf{Z}$, yielding an information matrix $\boldsymbol{\Lambda} (\mathbf{s})$ that varies with respect to the chosen sensor suite $\mathbf{s}$.

The objective becomes finding a suitable subset of sensors $\mathbf{s}$ such that the system's requirements are satisfactorily met. In order to maximize performance, it is desired to minimize the uncertainty of the SLAM solution, which is posed as
\begin{equation}\label{eq:logdet}
  \mathbf{s}^* = \arg\max_{\mathbf{s}\subseteq\mathcal{S}} \log \det (\boldsymbol{\Lambda} (\mathbf{s}))
\end{equation}
\noindent subject to the budget constraint $\text{Cost}(\mathbf{s}) = B$, where $B$
represents the available amount of money that can be spent on the sensor architecture.

\section{Sensor Selection Methodology}\label{sec:methodology}

In order to develop the proposed approach, a sensor library to act as the pool of available sensors $\mathcal{S}$ was crafted using data from real hardware. Due to the combinatorial nature of the problem, a greedy algorithm is chosen to direct the search within $\mathcal{S}$ for a near-optimal sensor combination.

\subsection{Sensor Library}

Since different types of sensors lead to different measurement noise characteristics, four types of sensors are considered herein: radars, LiDARs, stereo, and monocular cameras. To generate comparisons, a sensor library encompassing the original LiDAR and cameras used in the KITTI dataset~\cite{Geiger2013IJRR}, as well as commercially available sensors, is built; sensor details are presented in Table~\ref{StereoTable}.

\begin{table*}[h]
\caption{Attributes of considered sensors (*Field of View). Sensors with a start ($\star$) were used for the KITTI data collection. Other sensors have been added for consideration.}
\begin{center}
\begin{tabular}{|c|c|c|c|c|c|c|c|}
\hline
\textbf{Sensor} & \textbf{Range (m) } & \textbf{FOV* (rad)} & \textbf{$\sigma_{range}$ (m)} & \textbf{$\sigma_{bearing}$ (rad)} & \textbf{Cost (\$)} & \textbf{Data Source}\\
\hline
Long Range LiDAR $\star$ & 120 & 2$\pi$ & 0.084 & 0.00110 & 100,000 & \cite{Glennie2010} \\
\hline
Mid-Range Radar & 160 & 0.1 & 0.04 & 0.00175 & 2,830 & \cite{Bosch2019} \\
\hline 
Long-range Radar & 250 & 0.0698 & 0.013 & 0.00175 & 1,493 & \cite{Continental2017} \\
\hline
Mid-Range LiDAR & 100 & 2$\pi$ & 0.03 & 0.00524 & 4,000 & \cite{Velodyne2019} \\
\hline
\textbf{Stereo Camera} & \textbf{Range (m)} & \textbf{FOV *} (rad) & \textbf{Baseline (m)} & \textbf{Focal Length (pixels)} & \textbf{Cost (\$)} & \textbf{Data Source}\\
\hline
Wide angle stereo camera $\star$ & 50 & $\pi/2$ & 0.5371 & 721.5377 & 2,990 & \cite{Geiger2013IJRR}\\
\hline
HD2K stereo camera & 20 & 1.33 & 0.120 & 1400 & 449 & \cite{StereoLabs}\\
\hline
Low resolution stereo camera & 20 & 1.52 & 0.120 & 350 & 449 & \cite{StereoLabs}\\
\hline
\end{tabular}
\end{center}
\label{StereoTable}
\end{table*}

The output of all sensors is taken to be a range and bearing measurement to a landmark. Each sensor is characterized by their range and field of view (FOV) attributes, and a Gaussian noise model is assumed for their produced measurements. Manufacturer specifications are used to describe the sensors attributes, except for when a more detailed study is available. 

The main characteristics for each sensor type are:
\begin{itemize}
    \item {\bf LiDARs} have a 360 \degree FOV, with a long range. Empirical data is employed to choose noise characteristics for the long-range LiDAR~\cite{Glennie2010}, with the experimental Root Mean Square Error (RMSE) taken as the standard deviation for the noise model; bearing residuals were found to be lower than the manufacturer's specification.
    \item {\bf Radars} have the longest range, but a very narrow FOV of only several degrees. Their noise characteristics are comparable to the LiDAR's, but are much cheaper.
    \item {\bf Stereo cameras} have measurement noise models that depend on their intrinsic and extrinsic parameters. The standard deviation for bearing is taken to be the same as for mono cameras. An additional parameter to account for the depth-dependent range errors is utilized. We use the following accuracy formula as the standard deviation for range measurements:
\begin{equation} \label{eq:stereo}
    dz = \frac{z^2}{f \cdot b}\sigma_d
\end{equation}
where z is the distance to the object, f is the focal length, b is the stereo baseline and $\sigma_d$ is the pixel disparity error.
\end{itemize}

Sensor placement is also taken into account: LiDARs can be placed on top of the car, and the rest along 12 evenly spaced angles around the car. Therefore, with different models for each sensor type, a library with 62 elements is obtained ($5\times12$ for radar and cameras, plus two LiDARs). A measurement is obtained by one of the sensors if the corresponding landmark is within its sensing cone defined by the range and FOV.

\subsection{Sensor Architectures Enumeration and Evaluation}

With a cardinality of $N$ for the desired sensor suite size, an exhaustive comparison of architectures would need to evaluate $\binom{62}{N}$ possible combinations; for example, if $N=6$ is chosen, this yields a total of $6.15\times10^7$ possible architectures. Conversely, a greedy algorithm allows for the enumeration of only $ \sum_{i=57}^{62} i = 357$ architectures. The evaluation of each architecture is then done through a factor graph approach. A budget of $\$ 110,000$ is allocated for each sensor selection.

To prevent a full factorial enumeration, the greedy strategy dictates which sensor combination to evaluate next in the following manner: each entry $s_i$ in the library $\mathcal{S}$ is evaluated, and the item that led to the highest value of the objective function $J = \log \det (\boldsymbol{\Lambda} (s_i))$ is added to the sensor suite $\mathbf{s}$. Subsequently, each remaining item in the library is evaluated in conjunction with the previously chosen items -- that is $J = \log \det (\boldsymbol{\Lambda} (\mathbf{s}\cup s_i))$ -- to determine which one to pick next. The previous steps are then repeated until the cardinality constraint is satisfied. The strategy is summarized in Algorithm~\ref{alg:greedy}.

\begin{algorithm}[!htbp]                                                           
\caption{Greedy Enumeration Strategy}\label{alg:greedy}                            
\begin{algorithmic}[1]                                                             
    \Procedure{getGreedyArchitecture($\mathcal{S},B$)}{}             
        \State $\mathbf{s} \gets \{\}$    start with empty selection
        
        \While{$\text{Cost}(\mathbf{s}) < B$}
        \State $J_{max} \gets -\infty$
        \For{each $s_i \in {\mathcal{S}}$}          
            \State $J \gets \log\det(\boldsymbol{\Lambda}(\mathbf{s}\cup s_i))$
            \If{$J > J_{max}$}       
                \State $s^* \gets s_i$
                \State $J_{max} \gets J$
            \EndIf                                                    
        \EndFor
        \State $\mathbf{s} \gets \mathbf{s}\cup s^*$
        \State $\mathcal{S} \gets \mathcal{S}\setminus s^*$ 
    \EndWhile 
\State \Return $\mathbf{s}$                         
\EndProcedure                                                                   
\end{algorithmic}                                                               
\end{algorithm} 

\begin{figure}
\begin{subfigure}{.5\linewidth}
  \includegraphics[width=1\linewidth]{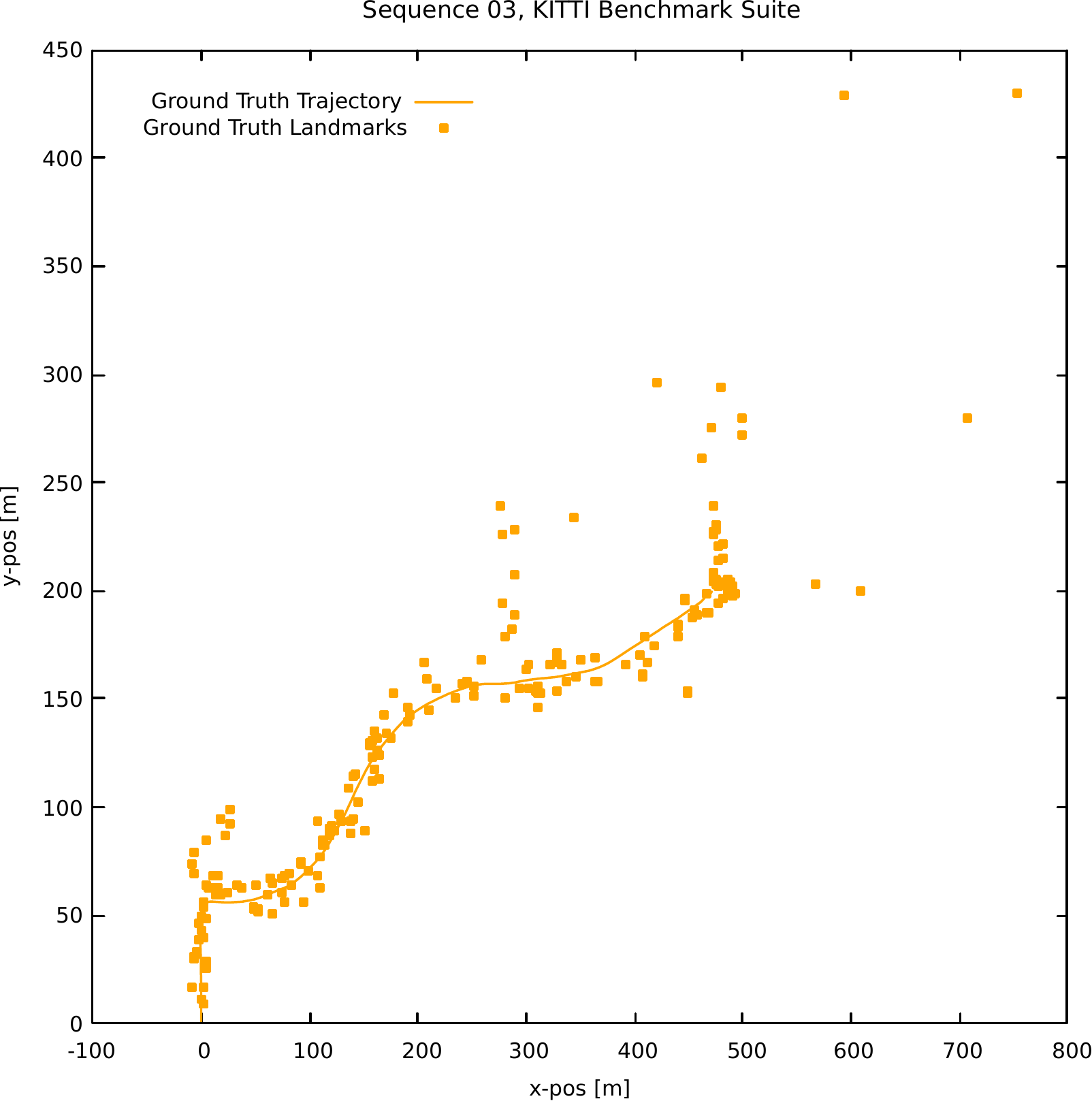}
\end{subfigure}%
\begin{subfigure}{.5\linewidth}
  \includegraphics[width=1\linewidth]{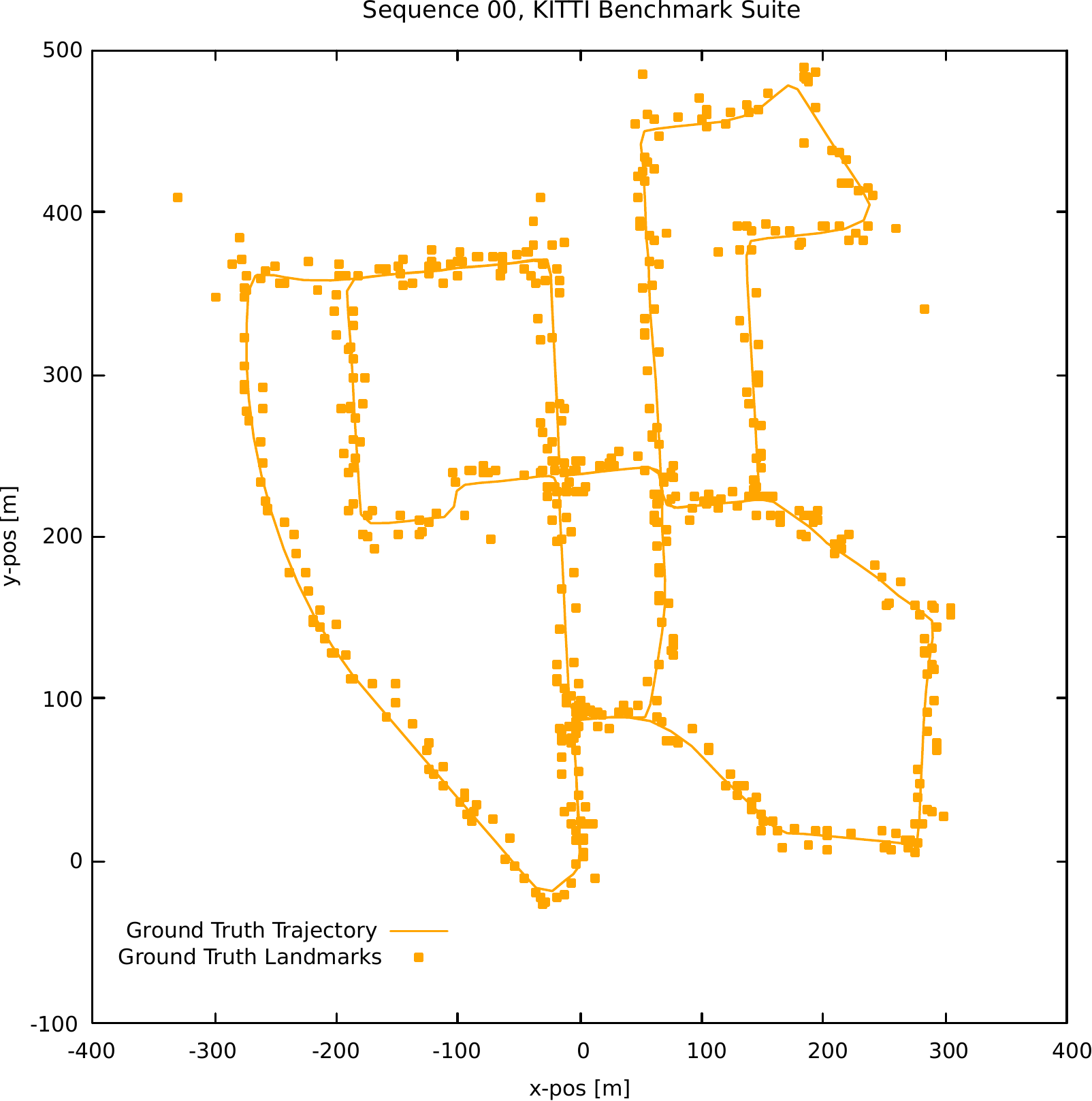}
\end{subfigure}
\caption{Route and landmarks for KITTI sequences 03 (left) and 00 (right). The axes represent the x and y coordinates of the car and landmark positions}
\label{fig:KITTIseq}
\end{figure}

For evaluation purposes, sequences 03 and 00 of the KITTI benchmark suite are employed. Sequence 03 consists of a relatively linear course resembling a typical suburban scenario. Conversely, sequence 00 contains many turns and loop closures, which resembles more closely operations within an urban scenario. The ground truth for both scenarios is shown in Figure~\ref{fig:KITTIseq}.

Using the methodology presented in this section, the following sensor recommendation is made to improve SLAM performance:
\begin{itemize}
    \item {\bf Greedy03}: long-range  LiDAR, long range radar at 12, 2, 6, and 11 o'clock,  wide-angle  stereo-cams at 6 o’clock, HD2K stereo-cam at 0 and 5 o’clock for sequence 03, with a cost of \$$109,860.00$.
    \item {\bf Greedy00}: long-range LiDAR,  HD2K stereo-cams at 6 and 12 o'clock, a mid-range radar at 12 o'clock, and long-range radars at 2, 7, 8, and 12 o'clock for sequence 00, with a cost of \$$109,700.00$.
\end{itemize}

These results show that the proposed methodology captures the impact of the environment on an ideal sensor combination. In sequence 03, most landmarks are situated in the front or back of the car, and at a close range, leading to the selection of only one LiDAR and stereo cameras facing the front and back of the car. With landmarks situated further away and the car experiencing more turns, in addition to the stereo cameras, the LiDAR is combined with radars facing front, right, back, and left for sequence 00.

Regardless of the performance level exhibited by these sensor combinations within their respective environments, there may be cheaper combinations that yield slightly higher, but still acceptable, levels of uncertainty in the MAP estimates. Additionally, if the long-range LiDAR fails for all or part of the trajectory, the performance greatly decreases in both sequences. Section \ref{sec:tradespace} elaborates on how this methodology can be adapted to understand the link between cost and SLAM performance, and Section \ref{sec:resilience} adds to the quality of the design by proposing resilient sensor selections. 

\section{Cost-Performance Tradespace Quantification}\label{sec:tradespace}

The proposed greedy algorithm yields one single architecture point; the one with the lowest uncertainty in the estimate. However, designers might be interested in understanding how additional investment in the sensor suite reduces this uncertainty. Hence, in order to generate more points in the tradespace, two modifications to the current strategy are made.     

Firstly, the performance of each intermediate combination below the cardinality constraint is recorded. Secondly, the objective function used by the greedy algorithm is adapted to take into account cost, by dividing performance by sensor cost, thus generating architectures at different cost levels. This way, several greedy algorithms are operated in parallel, using the following objectives:
\begin{itemize}
    \item uniform: $J = \log\det (\boldsymbol{\Lambda}(\mathbf{s}))$
    \item cost-benefit: $J = \log\det (\boldsymbol{\Lambda}(\mathbf{s})) / \text{Cost}(\mathbf{s})$
\end{itemize}

\begin{figure}[thpb]
  \centering
  \includegraphics[width=1\linewidth]{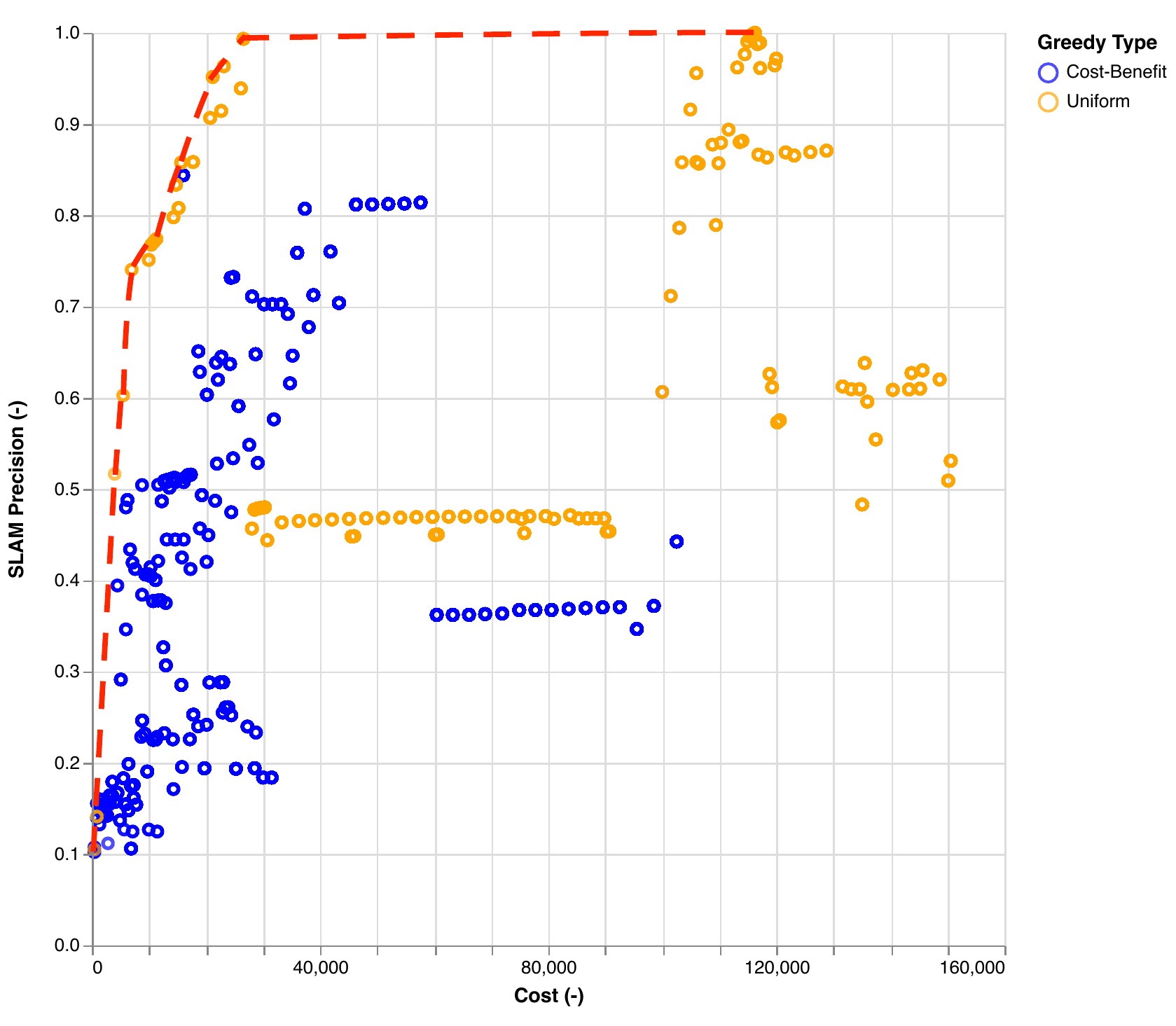}
  \caption{Pareto Front obtained by running different versions of the greedy algorithm on KITTI sequence 03. Performance is normalized. The dashed red line shows the Pareto Front. The orange architectures were generated by the uniform greedy algorithm. The cheapest architectures were generated by the cost-benefit algorithm, in blue.}
  \label{fig:ParetoFront}
\end{figure}

Figure \ref{fig:ParetoFront} shows the tradespace mapping results for sequence 03. The most expensive architectures cost more than \$100,000, and contain the long range LiDAR. However, there are architectures costing about \$27,000, which achieve about 99\% of the highest performance. These architectures contain mostly stereo cameras, and the cheaper LiDAR, meaning that with the sensor attributes chosen here, visual odometry combined with a higher noise LiDAR leads to a level of uncertainty almost as low as systems using expensive LiDARs, for a much lower hardware cost.

\section{Designing for Resilience}\label{sec:resilience}

Many adverse events can disrupt the correct operation of car sensors during operation. Hardware can permanently fail, or sensor data can become corrupted during certain periods of time, e.g., due to harsh reflections from buildings or mist for the LiDAR, or rainy scenes lowering the quality of a camera's vision output ~\cite{Gehrig2013}. Therefore, in order to select resilient sensor combinations, the following worst-case scenario is considered: by dividing the vehicle's route into smaller time periods, the sensor providing the most useful information at each particular time step is assumed to fail.

\begin{algorithm}[!htbp]                                                           
\caption{Resilient Greedy Enumeration Strategy}\label{alg:Robustgreedy}                            
\begin{algorithmic}[1]                                                             
     \Procedure{getResilientArchitecture($\mathcal{S},B$)}{}             
         \State $\mathbf{s_\text{res}} \gets \{\}$    start with empty selection
         \State $\mathbf{s_\text{fail}} \gets \{\}$    start with empty selection
        
         \State $s_{\text{best}} \gets \underset{e \in \mathcal{S}}{\arg\max} f(\{e\})$
        
         \State $\mathbf{s_\text{fail}} \gets s_{\text{best}}$
          \State $\mathcal{S} \gets \mathcal{S}\setminus s_{\text{best}}$
          \State $B \gets B-\text{Cost}(\mathbf{s_\text{fail}})$
        
         \State $\mathbf{s} \gets \textsc{getGreedyArchitecture}(\mathcal{S},B)$ 
        
         \State $\mathbf{s_\text{res}} \gets \mathbf{s_\text{fail}}\cup \mathbf{s}$
 \State \Return $\mathbf{s_\text{res}}$                         
 \EndProcedure                                                                   
\end{algorithmic}                                                               
\end{algorithm} 

The algorithm proposed by \cite{Tzoumas2018} is employed here within the autonomous vehicle design context and shown in Algorithm \ref{alg:Robustgreedy}. The strategy consists in, for each time period, evaluating each sensor individually using the problem's factor graph in order to identify the most critical sensor and its placement; it is the one assumed to fail. The rest of the selection is carried out through the proposed greedy algorithm, after the critical sensor $\mathbf{s}_+$ has been removed from the library of available components ($\mathcal{S} \setminus \mathbf{s}_+$). The allocated budget remains \$110,000 for the total sensor selection, including the failing sensor.
The  method is ran for every time period in the sequence; in this case, a time period corresponds to several seconds. Because of this assumption, the selection is now performed online, meaning that sensors can be turned on or off along the route.

\begin{figure*}
    \centering
    \includegraphics[width=\textwidth]{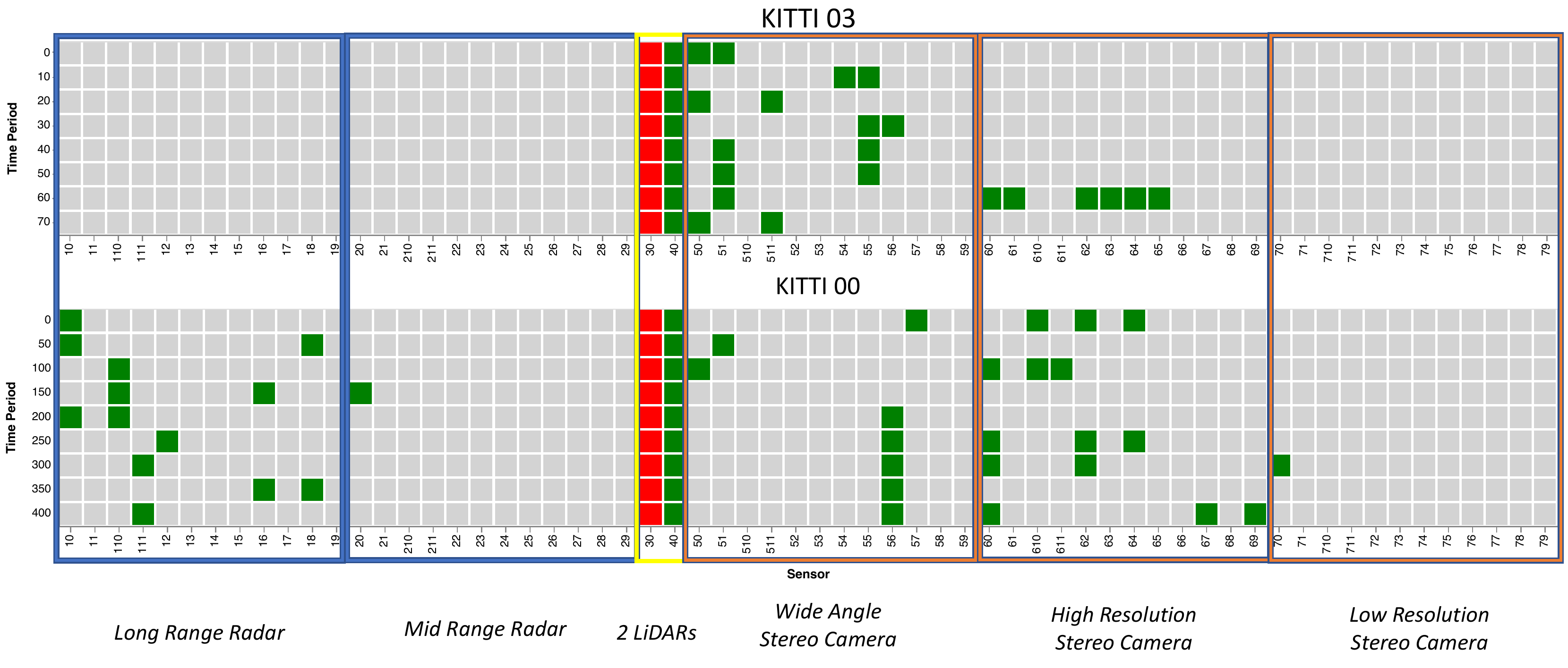}
    \caption{Sensor choice over time for KITTI sequences 03 (top) and 00 (bottom). The rows represent the different time periods, and the columns are the different sensor indices; first index denotes the sensor type, with the remaining digits denoting its placement with respect to the vehicle's frame. Sensors providing the most information at each time period are assumed to fail and are denoted in red. Selected active sensors are green.}
    \label{fig:onlineRobust}
    \vspace{-5mm}
\end{figure*}

Figure~\ref{fig:onlineRobust} shows the sensor selection for each time period, with the failing sensors marked as red, and active sensors in green. The selection for the sequence 03 -- represented in the upper grid of the figure -- is firstly analyzed. Each block represents a sensor type, and within each block are sensor placements. For example, 26 represents the mid-range radar placed facing 6 o'clock, the back of the car.

As per the algorithm's choice, the long range LiDAR, the sensor that would have provided the most information throughout the trajectory, is designated as inoperative. To compensate for this loss, the mid range LiDAR is now part of the selection for every time period. With both LiDARs selected, only \$6,000 remain in the budget. For most time periods, they are consumed by two wide-angle stereo cameras placed alternatively in the front or in the back of the car, or for one time period, many HD stereo cameras facing different directions on the right side of the car. The front-back axis remains salient in this selection. Whereas the value of our performance metric for the non resilient selection is 865.40 if the long-range LiDAR fails, the robust selection obtains a score of 1011.58; a performance improvement of about 17 \%.

With the same budget, designing for resilience, rather than for performance in {\bf Greedy03}, therefore leads to selecting the second LiDAR and the absence of radars.



For the longer KITTI sequence 00 -- shown in the lower grid of figure \ref{fig:onlineRobust} -- the most useful sensor is also the long-range LiDAR, and the same strategy for resilience emerges; the mid-range LiDAR is selected at every time period. However, the difference between resilient selections for both environment lies in the use of radars; in KITTI 00, the long-range radar is employed in many different directions. As it is twice cheaper than the wide-angle stereo camera, more HD stereo cameras can be used, also in various directions. The front-back preference does not appear here, which is not surprising, as the track of sequence 00 provides much more landmarks in different directions, whereas 03 is more unidirectional. Because landmarks are, on average, further away from the car, the long-range radar provides more information than the wide-angle stereo camera in this instance.

\begin{figure}
\vspace{-7mm}
\begin{subfigure}{\linewidth}
    \vspace{1cm}
  \includegraphics[width=\linewidth]{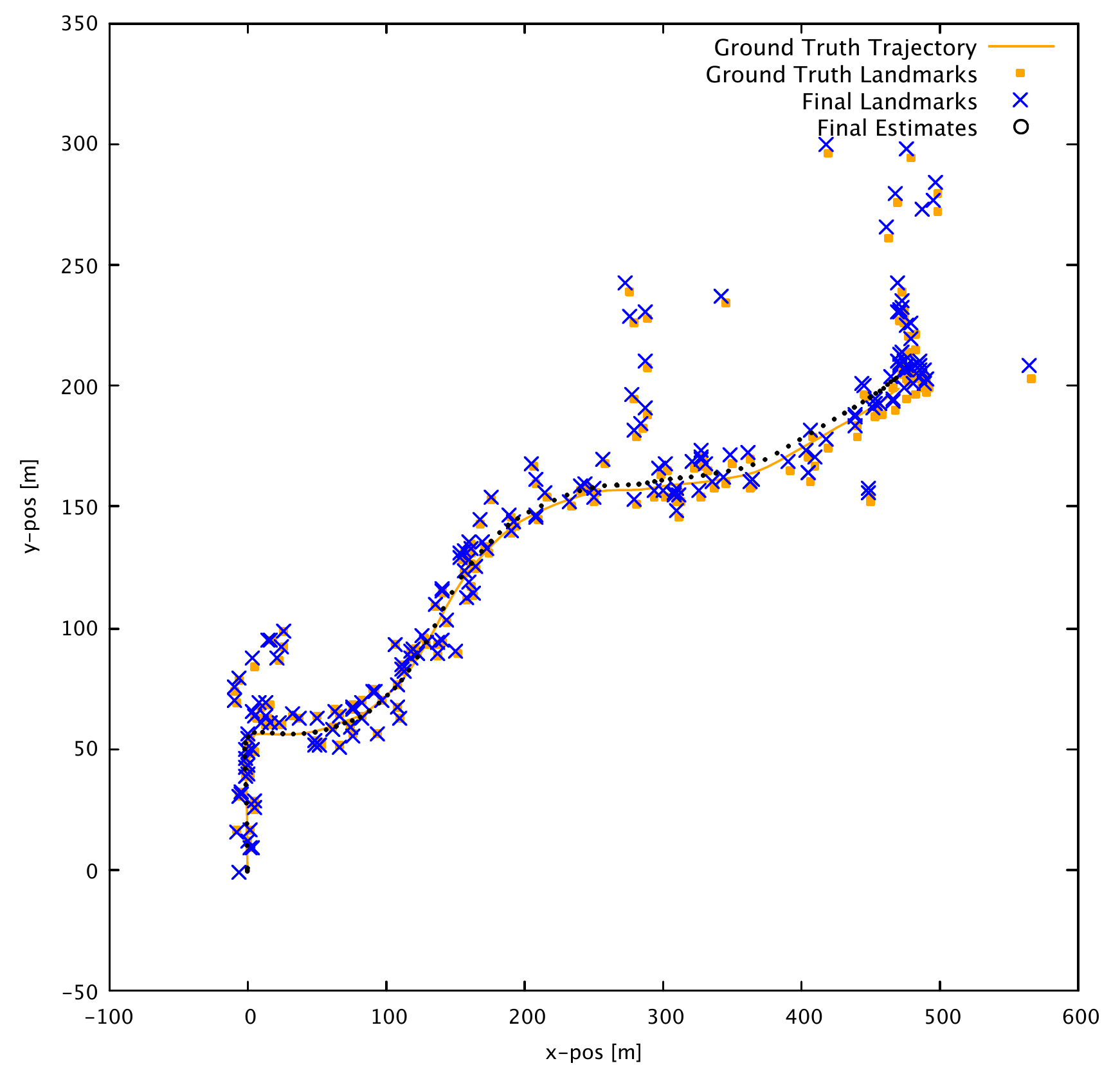}
\end{subfigure}
\par\bigskip 
\begin{subfigure}{\linewidth}
  \includegraphics[width=\linewidth]{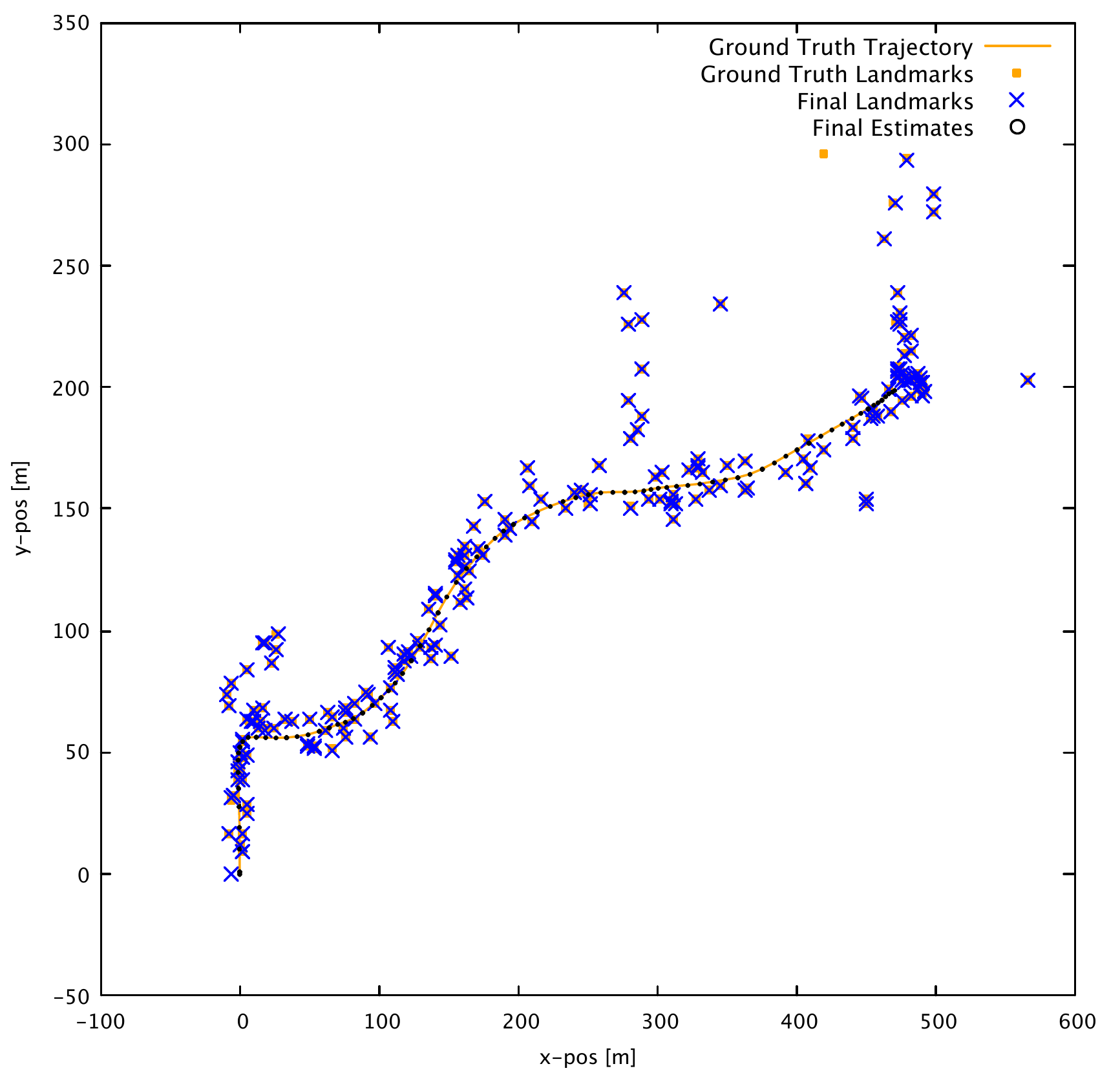}
\end{subfigure}
\caption{Accuracy comparison of {\bf Greedy03} and {\bf Resilient03} sensor architectures for KITTI 03 sequence: non-resilient uniform cost shown on the top with one sensor failing, versus resilient with one sensor failing on the bottom.}
\label{fig:SelectionComparison}
\vspace{-7mm}
\end{figure}

A comparison of the final SLAM performance for the worst-case scenarios of {\bf Greedy03} and the {\bf Resilient03} is shown in Figure~\ref{fig:SelectionComparison}. The estimated trajectory and landmark positions are displayed, which portray the robustness levels of both sensing architectures. In the non resilient case, the estimation error grows, reaching more than 5 meters per position and landmark, whereas the resilient combination provides an accurate estimate throughout the trajectory. 

This section shows that different types of sensors have to be installed on the vehicle to guarantee an acceptable level of performance in case of failures or adverse scenarios. For those settings, a LiDAR with higher noise levels and a shorter range is chosen to compensate for the best LiDAR, combined with stereo cameras, especially facing directions towards where most landmarks are located, and radars when landmarks are located far away from the car.

\section{Conclusion}\label{sec:conclusion}

Several lessons can be drawn from the different experiments presented in this paper. Firstly, if the landmark density is predominantly at close ranges with respect to the vehicle, a combination of strategically placed stereo cameras with a low-cost LiDAR can provide as low of a SLAM uncertainty as combinations of fewer sensors that involve the use of a high-end LiDAR. Secondly, landmark positions in the environment relative to the car, as well as their number, influence the type of sensor choice and its placement within the vehicle. 

The resulting added value for resiliency of complementary sensor types emerges as an output of the proposed methodology, and is quantified for the first time. It is found that the design of a resilient sensor selection warrants a different sensor set, but the quantification of resiliency enables designers to make informed decisions on the sensor choices.

This paper provides a sensor selection recommendation in a widely used setting, with realistic sensor attributes, considering different figures of merit for the design; SLAM performance, cost, and resiliency. Future work involves including data association risks when considering sensors, as well as translating driving practices or regulations in design requirements for the uncertainty in SLAM systems, in order to understand which regions of the Pareto front are acceptable.









\bibliographystyle{ieeetr}
\bibliography{AeroAstro-Recherche-IROSbibli.bib}

\begin{thebibliography}{10}

\bibitem{Junietz2018}
P.~Junietz, F.~Bonakdar, B.~Klamann, and H.~Winner, ``{Criticality Metric for
  the Safety Validation of Automated Driving using Model Predictive Trajectory
  Optimization},'' in {\em IEEE Conference on Intelligent Transportation
  Systems, Proceedings, ITSC}, vol.~2018-Novem, pp.~60--65, 2018.

\bibitem{Chong2015}
T.~J. Chong, X.~J. Tang, C.~H. Leng, M.~Yogeswaran, O.~E. Ng, and Y.~Z. Chong,
  ``{Sensor Technologies and Simultaneous Localization and Mapping (SLAM)},''
  {\em Procedia Computer Science}, vol.~76, no.~Iris, pp.~174--179, 2015.

\bibitem{Broggi2013}
A.~Broggi, M.~Buzzoni, S.~Debattisti, P.~Grisleri, M.~C. Laghi, P.~Medici, and
  P.~Versari, ``Extensive tests of autonomous driving technologies,'' {\em IEEE
  Transactions on Intelligent Transportation Systems}, vol.~14, no.~3,
  pp.~1403--1415, 2013.

\bibitem{Ulbrich2017}
S.~Ulbrich, A.~Reschka, J.~Rieken, S.~Ernst, G.~Bagschik, F.~Dierkes, M.~Nolte,
  and M.~Maurer, ``{Towards a Functional System Architecture for Automated
  Vehicles}.'' 2017.

\bibitem{ISO26262}
{International Standardization Organization}, ``{ISO26262-2 - Road vehicles -
  Functional safety - Part 2: Management of functional safety},'' tech. rep.,
  International Standardization Organization, 2018.

\bibitem{Bosch2018}
P.~M. B{\"{o}}sch, F.~Becker, H.~Becker, and K.~W. Axhausen, ``{Cost-based
  analysis of autonomous mobility services},'' {\em Transport Policy}, vol.~64,
  pp.~76--91, may 2018.

\bibitem{Ort2018}
T.~Ort, L.~Paull, and D.~Rus, ``{Autonomous Vehicle Navigation in Rural
  Environments without Detailed Prior Maps},'' in {\em International Conference
  on Robotics and Automation}, 2018.

\bibitem{Jo2015}
K.~Jo, Y.~Jo, J.~K. Suhr, H.~G. Jung, and S.~Member, ``{Precise Localization of
  an Autonomous Car Based on Probabilistic Noise Models of Road Surface Marker
  Features Using Multiple Cameras},'' {\em IEEE Transactions on Intelligent
  Transportation Systems}, vol.~16, no.~6, pp.~3377--3392, 2015.

\bibitem{Wadud2017}
Z.~Wadud, ``{Fully automated vehicles: A cost of ownership analysis to inform
  early adoption},'' {\em Transportation Research Part A: Policy and Practice},
  vol.~101, pp.~163--176, jul 2017.

\bibitem{Tzoumas2017}
V.~Tzoumas, L.~Carlone, G.~J. Pappas, and A.~Jadbabaie, ``{Sensing-Constrained
  LQG Control},'' in {\em American Control Conference (ACC)}, p.~14, sep 2018.

\bibitem{Zhao2018}
Y.~Zhao and P.~A. Vela, ``{Good Feature Selection for Least Squares Pose
  Optimization in VO/VSLAM},'' in {\em 2018 IEEE/RSJ International Conference
  on Intelligent Robots and Systems (IROS)}, pp.~1183--1189, IEEE, oct 2018.

\bibitem{carlone2018attention}
L.~Carlone and S.~Karaman, ``{Attention and anticipation in fast
  visual-inertial navigation},'' {\em IEEE Transactions on Robotics}, 2018.

\bibitem{Collin2018}
A.~Collin and A.~{Teran Espinoza}, ``{SLAM-Based Performance Quantification of
  Sensing Architectures for Autonomous Vehicles},'' in {\em 2018 IEEE
  International Conference on Vehicular Electronics and Safety (ICVES) (ICVES
  2018)}, (Madrid, Spain), sep 2018.

\bibitem{Dellaert2012}
F.~Dellaert, ``{Factor Graphs and GTSAM : A Hands-on Introduction},'' Tech.
  Rep. September, GeorgiaTech, Atlanta, 2012.

\bibitem{kaess2008isam}
M.~Kaess, A.~Ranganathan, and F.~Dellaert, ``isam: Incremental smoothing and
  mapping,'' {\em IEEE Transactions on Robotics}, vol.~24, no.~6,
  pp.~1365--1378, 2008.

\bibitem{Kaess2009}
M.~Kaess and F.~Dellaert, ``{Covariance recovery from a square root information
  matrix for data association},'' {\em Robotics and Autonomous Systems},
  vol.~57, pp.~1198--1210, dec 2009.

\bibitem{Krause2014}
A.~Krause and D.~Golovin, ``{Submodular Function Maximization},'' 2014.

\bibitem{joshi2009sensor}
S.~Joshi and S.~Boyd, ``{Sensor Selection via Convex Optimization},'' {\em IEEE
  Transactions on Signal Processing}, vol.~57, pp.~451--462, feb 2009.

\bibitem{Krause2008}
A.~Krause, J.~Leskovec, C.~Guestrin, J.~VanBriesen, and C.~Faloutsos,
  ``Efficient sensor placement optimization for securing large water
  distribution networks,'' {\em Journal of Water Resources Planning and
  Management}, vol.~134, no.~6, pp.~516--526, 2008.

\bibitem{Carrillo2012}
H.~Carrillo, I.~Reid, and J.~A. Castellanos, ``{On the comparison of
  uncertainty criteria for active SLAM},'' in {\em IEEE International
  Conference on Robotics and Automation}, pp.~2080--2087, 2012.

\bibitem{Geiger2013IJRR}
A.~Geiger, P.~Lenz, C.~Stiller, and R.~Urtasun, ``{Vision meets Robotics: The
  KITTI Dataset},'' {\em International Journal of Robotics Research (IJRR)},
  2013.

\bibitem{Glennie2010}
C.~Glennie and D.~D. Lichti, ``{Static Calibration and Analysis of the Velodyne
  HDL-64E S2 for High Accuracy Mobile Scanning},'' {\em Remote Sensing},
  vol.~2, pp.~1610--1624, jun 2010.

\bibitem{Bosch2019}
Bosch, ``{Mid-range Radar Sensor (MRR)},'' 2019.

\bibitem{Continental2017}
Continental, ``{ARS 408-21 Premium Long Range Radar Sensor 77 GHz},'' tech.
  rep., 2017.

\bibitem{Velodyne2019}
Velodyne, ``{Velodyne LiDAR - Puck},'' 2019.

\bibitem{StereoLabs}
StereoLabs, ``{What is the camera focal length and field of view?}.''

\bibitem{Gehrig2013}
S.~Gehrig, M.~Reznitskii, N.~Schneider, U.~Franke, and J.~Weickert, ``Priors
  for stereo vision under adverse weather conditions,'' in {\em Proceedings of
  the IEEE International Conference on Computer Vision Workshops},
  pp.~238--245, 2013.

\bibitem{Tzoumas2018}
V.~Tzoumas, A.~Jadbabaie, and G.~J. Pappas, ``{Resilient Non-Submodular
  Maximization over Matroid Constraints},'' {\em arXiv preprint
  arXiv:1804.01013}, 2018.

\end{thebibliography}

\end{document}